\documentclass[conference]{IEEEtran}
\IEEEoverridecommandlockouts

\usepackage[colorlinks=true, allcolors=blue]{hyperref}
\usepackage{cite}
\usepackage{amsmath,amssymb,amsfonts}
\usepackage{algorithmic}
\usepackage{graphicx}
\usepackage{textcomp}
\usepackage{xcolor}
\usepackage{gensymb}
\usepackage{tabu}
\usepackage{booktabs}
\def\BibTeX{{\rm B\kern-.05em{\sc i\kern-.025em b}\kern-.08em
    T\kern-.1667em\lower.7ex\hbox{E}\kern-.125emX}}
\begin{document}

\title{In-Bed Pose Estimation: A Review}

\author{\IEEEauthorblockN{Ziya Ata Yazıcı}
\IEEEauthorblockA{\textit{Istanbul Technical University} \\
Istanbul, Turkey \\
yaziciz21@itu.edu.tr}
\and
\IEEEauthorblockN{Sara Colantonio}
\IEEEauthorblockA{\textit{ISTI-CNR} \\
Pisa, Italy \\
sara.colantonio@isti.cnr.it}
\and
\IEEEauthorblockN{Hazım Kemal Ekenel}
\IEEEauthorblockA{\textit{Istanbul Technical University},
Istanbul, Turkey \\
\textit{Qatar University}, Doha, Qatar \\
ekenel@itu.edu.tr,  hekenel@qu.edu.qa}}

\maketitle

\begin{abstract}
Human pose estimation, the process of identifying joint positions in a person's body from images or videos, represents a widely utilized technology across diverse fields, including healthcare. One such healthcare application involves in-bed pose estimation, where the body pose of an individual lying under a blanket is analyzed.
This task, for instance, can be used to monitor a person's sleep behavior and detect symptoms early for potential disease diagnosis in homes and hospitals. Several studies have utilized unimodal and multimodal methods to estimate in-bed human poses. The unimodal studies generally employ RGB images, whereas the multimodal studies use modalities including RGB, long-wavelength infrared, pressure map, and depth map. Multimodal studies have the advantage of using modalities in addition to RGB that might capture information useful to cope with occlusions. Moreover, some multimodal studies exclude RGB and, this way, better suit privacy preservation.
To expedite advancements in this domain, we conduct a review of existing datasets and approaches. Our objectives are to show the limitations of the previous studies, current challenges, and provide insights for future works on the in-bed human pose estimation field. 

\end{abstract}


\section{Introduction}
Human pose estimation involves predicting the joints, e.g., head, elbow, and knee, of a human body from an image or video. Since human pose serves as a key technology for human perception, it has been utilized in a wide range of applications~\cite{wang2021deep}. The medical field has also benefited from human pose analysis by continuously monitoring individuals in home and hospital environments. Examples of practical applications include assistive systems for elderly individuals\cite{jang2020etri}, estimating infant poses\cite{groos2022towards}, automatic rehabilitation systems\cite{qiu2022pose}, and tracking the poses of surgeons and clinicians\cite{srivastav2022unsupervised}. In addition to the given topics, an emerging research area, in-bed human pose estimation has also received attention recently. This topic involves tracking individuals' poses while lying or sleeping in a bed. These systems can provide intuitive insights into diagnosing disorders by observing the symptoms during sleep, evaluating sleep quality, and monitoring the sleeping position changes to improve post-surgery healing. However, occlusion brings a challenge to the task when the person is covered with a blanket. Hence, the available data modalities should be utilized to see under the occlusion.

To accelerate the studies on in-bed pose estimation and showcase the current limitations and potentials, our study will be the first review paper that covers the available datasets, used metrics, and the previous studies specific to the in-bed pose estimation topic. First, we will review the available datasets, the number of samples they include, the modalities they are composed of, and the evaluation metrics used to evaluate the joint predictions. Second, we will categorize the in-bed pose estimation works into two groups: unimodal and multimodal methods. Unimodal studies use only a single modality to estimate human pose, while multimodal studies use multiple modalities to benefit from color, heat, pressure,  and depth maps of the scene. Additionally, since the main application areas of such systems are the hospital rooms and intensive care units, one concern of the approaches is the preservation of the patient's privacy. Thus, besides RGB, the use of long-wavelength infrared (LWIR), depth maps, and pressure maps have been mainly used in the literature. Finally, the current limitations and directions of in-bed pose estimation will be given.

 \begin{figure}
\centering
\includegraphics[scale=0.2]{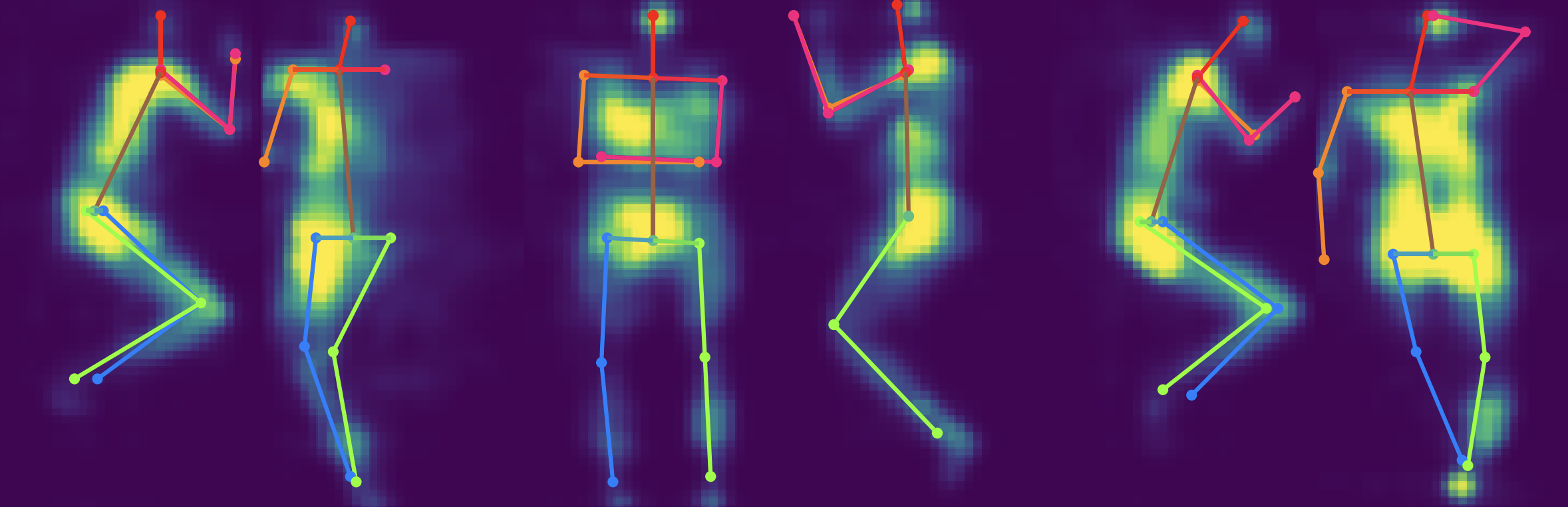}
\caption{Sample pressure maps and estimated poses from the Pressure-Sensing Mat Dataset \cite{clever20183d}.} \label{fig:PM}
\end{figure}

\section{Datasets}

\begin{table*}
\centering
\caption{A summary of the publicly available datasets on in-bed pose estimation}
\label{tab:dataset}
\resizebox{\textwidth}{!}{%
\begin{tabular}{llllll} 
\hline
\textbf{Dataset} & \textbf{Modality} & \textbf{\# Participants} & \textbf{\# Samples} & \textbf{Labels} & \textbf{Type} \\ 
\hline\hline
Pressure-Sensing Mat Dataset\cite{clever20183d} & Pressure Map & 17 & 28,000 images & 3D Pose & Image \\
Mannequin In-Bed Dataset\cite{liu2019bed} & RGB, LWIR & 2 & 419 images & 2D Pose & Image \\
BlanketSet\cite{carmona2022blanketset} & RGB, LWIR, Depth Map & 14 & 303,965 frames & Action Class & Video \\
SLP Dataset\cite{liu2022simultaneously} & RGB, LWIR, Depth Map, Pressure Map & 109 & 14,715 frames & 3D Pose& Video \\
Patient MoCap Dataset\cite{achilles2016patient} & RGB, Depth Map & 10 & 180,000 frames & 3D Pose& Video \\
\hline
\end{tabular}
}
\end{table*}

In this section, we review the publicly available datasets collected for in-bed pose estimation tasks. These encompass various patient poses, lighting conditions, and occlusion scenarios. A summary of the datasets is also given in Table~\ref{tab:dataset}.

\subsubsection{\textbf{Pressure-Sensing Mat Dataset\cite{clever20183d}}} This is a public unimodal pose estimation dataset with a pressure map of 17 human participants in a hospital bed. The dataset includes over 28,000 pressure maps with 3D human pose annotations in a 17-joint skeleton model. The dataset includes two configurations, e.g., supine (0\textdegree) and seated (60\textdegree), and the participants took various poses during the data collection. The goal of the dataset is to enable patient-assisting robots to estimate the 3D position of the patient while minimizing the impact of lighting and occlusion on the bed. Sample images and annotations from the dataset can be seen in Figure \ref{fig:PM}.

 \begin{figure}[h]
\centering
     \includegraphics[scale=0.5]{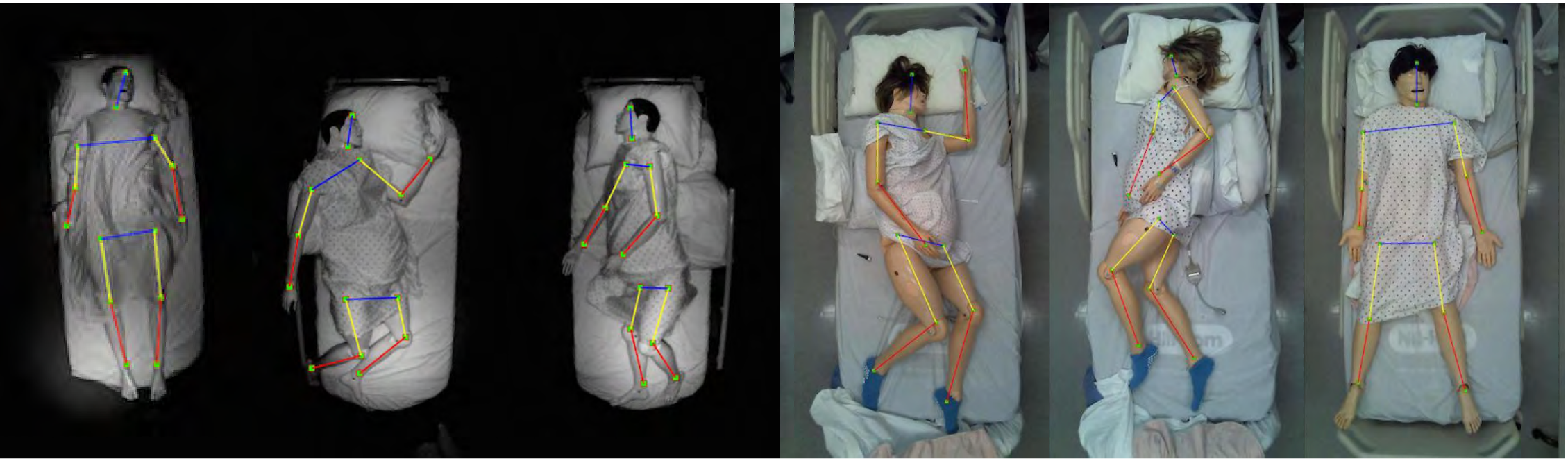}
\caption{Samples images from the Mannequin In-Bed Dataset\cite{liu2019bed} in two modalities: LWIR and RGB modalities from left to right.} \label{fig:Mannequin}
\end{figure}

\subsubsection{\textbf{Mannequin In-Bed Dataset \cite{liu2019bed}}} This is the first public dataset composed for in-bed pose estimation studies, collected by utilizing realistic male and female mannequins in different poses, e.g., supine, left-side lying, and right-side lying, in a hospital room setting. Images are taken with an infrared selective (IRS) image acquisition system, and two sets of normal and infrared-illuminated images are included in the dataset. The poses of the images were annotated in a 14-joint format, and a total of 419 non-occluded pose images with 2D ground-truth skeletons are available in the dataset. Sample images and annotations from the dataset can be seen in Figure~\ref{fig:Mannequin}.

\subsubsection{\textbf{BlanketSet Dataset \cite{carmona2022blanketset}}} This is a public multimodal video-based action recognition dataset with RGB, LWIR, and depth map modalities recorded from 14 participants in eight movement sequences: Foot-to-knee, knee bend, swinging legs, hands to shoulders, belly-down spread, torso lean, stretched arms and feet stretched. The videos were recorded with three different blanket positions and blanket types, e.g., thickness, color, and weight. In total, 405 videos were collected.

 \begin{figure}[!h]
\centering
     \includegraphics[scale=0.33]{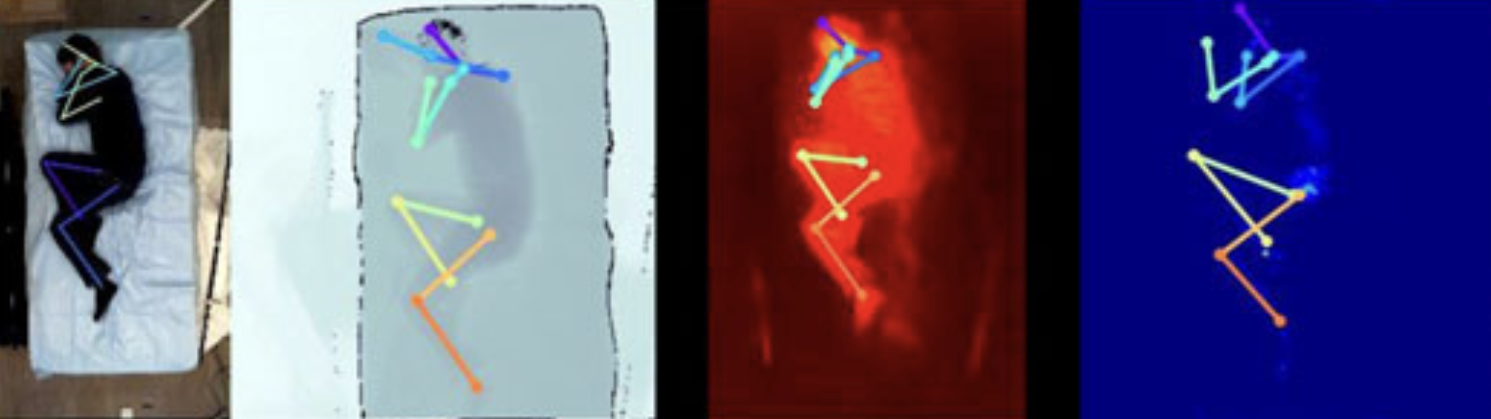}
\caption{Samples images from the SLP dataset\cite{liu2022simultaneously} in four modalities: RGB, depth map, LWIR, and pressure map from left to right.} \label{fig:SLP}
\end{figure}

\subsubsection{\textbf{Simultaneously-Collected Multimodal Lying
Pose (SLP) Dataset \cite{liu2022simultaneously}}} This is a large-scale public multimodal pose estimation dataset with 109 subjects and 14,715 samples in RGB, LWIR, depth, and pressure map modalities. Occlusion cases with different blanket thicknesses were also included. The dataset is divided into two parts: In-home (102 participants) and hospital settings (7 participants). From the participants, 15 poses were collected in three different positions; supine, left, and right-side sleeping were collected, and annotated in 14 joints. Sample images and annotations from the dataset can be seen in Figure \ref{fig:SLP}.

 \begin{figure}[h]
\centering
     \includegraphics[scale=0.37]{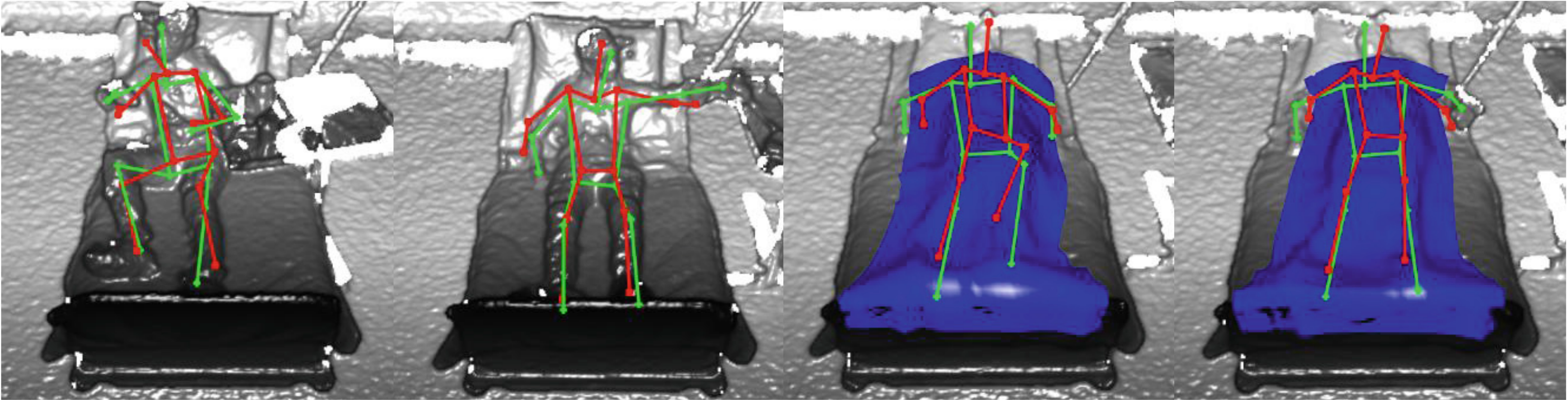}
\caption{Non-occluded and occluded sample depth maps from the Patient MoCap Dataset\cite{achilles2016patient}.} \label{fig:MoCap}
\end{figure}

\subsubsection{\textbf{Patient MoCap Dataset \cite{achilles2016patient}}} This dataset comprises sequences in RGB and depth map, each with ground truth pose information obtained from five calibrated motion capture cameras. These cameras track 14 rigid targets on each subject, allowing for the inference of 14 body joint locations. The subjects, consisting of five females and five males, performed ten sequences involving various activities such as getting in/out of bed, sleeping on horizontal/elevated beds, eating with/without clutter, using objects, reading, clonic movements simulating epileptic seizures, and a calibration sequence. The dataset includes 180,000 video frames captured by a calibrated and synchronized RGB-D sensor. Sample images and annotations from the dataset can be seen in Figure \ref{fig:MoCap}.

\section{Methods}

This section includes the unimodal and multimodal approaches used for the in-bed pose estimation task. Each study will be investigated by the utilized modalities, selected models, and the proposed approach. A summary of the approaches is given in Table~\ref{tab:methods}.

\begin{table*}
\centering
\caption{A summary of the proposed approaches on in-bed pose estimation}
\label{tab:methods}
\resizebox{\linewidth}{!}{%
\begin{tabu}{lllll} 
\toprule
\textbf{Studies} & \textbf{Highlight} & \textbf{Dataset} & \textbf{Metric} & \textbf{Modality} \\ 
\hline\hline
Liu and Ostadabbas\cite{liu2017vision} & A person-independent, latent variable-based hierarchical inference model was designed. & Mannequin\cite{liu2019bed} & Accuracy & RGB \\
Liu \textit{et al.}\cite{liu2019bed} & A CNN-based pose estimation with HOG rectification in non-occluded scenarios was performed. & Mannequin\cite{liu2019bed} & PCK & LWIR \\
Davoodnia \textit{et al.}\cite{davoodnia2022estimating} & An autoencoder model was used to change the pressure map representation for pose estimation. & Pressure\cite{clever20183d} & MPJPE & Pressure\\
Bigalke \textit{et al.}\cite{bigalke2023anatomy} & Anatomical loss-based unsupervised and source-free domain adaptation methods were implemented. & SLP\cite{liu2022simultaneously} & MPJPE & Depth \\
Afham \textit{et al.}\cite{afham2022towards} & Two GAN-based augmentation methods were used to increase the labeled images with occlusion. & SLP\cite{liu2022simultaneously} & PCK & LWIR \\
Casas \textit{et al.}\cite{casas2019patient} & Hash and CNN-based two approaches were implemented for pose estimation on pressure maps. & \textit{Private} & MAE & Pressure \\
Obeidavi \textit{et al.}\cite{obeidavi2022pose} & A multi-scale CNN model was proposed for thermal image-based pose estimation. & SLP\cite{liu2022simultaneously} & PCK & LWIR \\
Cao \textit{et al}.\cite{cao2022bed} & A variational autoencoder is designed to retrieve the missing RGB modality during inference time. & SLP\cite{liu2022simultaneously} & PCK & RGB, LWIR \\
Dayarathna \textit{et al}.\cite{dayarathna2023privacy} & Different modality fusion approaches and best performing modality couple were investigated. & SLP\cite{liu2022simultaneously} & PCK & RGB, LWIR, Depth, Pressure \\
Yin \textit{et al}.\cite{yin2022multimodal} & A modality fusion pyramid with an attention-based reconstruction method was designed. & SLP\cite{liu2022simultaneously} & MPJPE & RGB, LWIR, Depth, Pressure \\
\bottomrule
\end{tabu}
}
\end{table*}

\subsection{Unimodal In-bed Pose Estimation}

Several studies utilize a single source of information for in-bed pose estimation. In \cite{liu2017vision}, as one of the earliest approaches to in-bed pose estimation topic, the authors introduce a vision-based tracking system designed for long-term monitoring of in-bed postures in various environments. The system uses a latent variable-based hierarchical inference model to generate in-bed posture tracking history reports from top-view videos captured by regular off-the-shelf cameras. Despite being supervised, the model is person-independent and can be applied to new users without retraining. Experiments are performed on the mannequin dataset and a private human dataset with 358 samples from 12 participants. The results demonstrate the posture detection accuracy as 91.0\% on the mannequins dataset \cite{liu2019bed} and 93.6\%  on the private dataset with human participants.

In \cite{liu2019bed}, the authors propose a pre-trained Convolutional Neural Network (CNN) model, named Convolutional Pose Machine (CPM) to estimate in-bed poses on their composed Mannequin In-Bed Dataset. While collecting the images, the challenges unique to in-bed poses, such as lighting variations and unconventional perspectives, are addressed through an infrared selective (IRS) image acquisition technique. The collected images are processed with a Histogram of Oriented Gradient (HOG) rectification method to handle unusual positions and angles of the subject. They fine-tune the intermediate layers of the model stages with limited data and achieve an 82.6\% probability of correct
keypoint (PCK) with a threshold of 20\%, a metric for joint correctness based on the distance between the predicted and ground truth joints compared to a percentage of a person's torso length.

In \cite{davoodnia2022estimating}, the authors explore the application of in-bed pose estimation using pressure data. The study evaluates various strategies for detecting body pose from ambiguous pressure data by leveraging pre-existing pose estimators. The approaches include using pose estimators pre-trained on the RGB domain, re-training them on pressure datasets, and employing a learnable pre-processing domain adaptation step to transform pressure maps into a new representation. The authors introduce a domain adaptation method with a fully convolutional network, PolishNetU, to generate robust representations for common pre-trained pose estimation models. PolishNetU incorporates an objective function addressing pose identification loss, reconstructing the lost body parts, and penalizing large deviations from the original pressure maps. On the Pressure-Sensing Mat Dataset\cite{clever20183d}, the authors re-train pre-existing image-based pose estimators using the new representations of the pressure maps, which significantly increases the pose estimation accuracy compared to using the pressure maps directly.

In \cite{bigalke2023anatomy}, the authors propose a novel domain adaptation method that adapts a labeled source model to an unlabeled target domain, focusing on in-bed pose estimation. Two complementary adaptation strategies based on prior knowledge about human anatomy are introduced. The first strategy -- Unsupervised Domain Adaptation (UDA) -- guides the learning process in a supervised manner for the source domain and an unsupervised manner for the target domain, achieved through embedding anatomical constraints into an anatomical loss function. The second strategy -- Source-Free Domain Adaptation (SFDA) -- involves filtering pseudo labels for self-training based on their anatomical plausibility. The evaluation is conducted under two adaptation scenarios using the depth maps from the SLP~\cite{liu2022simultaneously} dataset and a newly created dataset from 13 participants. The proposed method reduces the domain gap and surpasses the baseline model by a 95.5 mean per joint position error (MPJPE), calculated as the average L2 Norm of differences between predicted and ground truth joint coordinates.

In \cite{afham2022towards}, the researchers introduce a methodology for cross-domain in-bed pose estimation, particularly in the LWIR modality. The formulation of the problem involves standard 2D human pose estimation tasks with labeled source domain data and unlabeled target domain data. To mitigate the domain gap, based on the work of Zhu \textit{et al.} \cite{zhu2017unpaired}, the authors propose a two-fold data augmentation pipeline, incorporating CycAug for unpaired image-to-image translation and ExtremeAug for introducing more covering artifacts and occlusions. The augmented images are then utilized for pose estimation, extending the input space for optimization. Furthermore, knowledge distillation is employed to transfer knowledge from a teacher to a student model, enhancing student performance. On the SLP dataset, by training only with the uncovered images and performing domain transfer into covered images, the approach achieves a 76.13\% PCK0.5 score.

In \cite{casas2019patient},  the objective is to explore in-bed motion monitoring by utilizing pressure sensors as a privacy-preserving alternative to video-based methods. Two approaches are presented: a hashing content-retrieval approach and a deep learning method for human pose estimation from pressure sensor data. The hashing-based approach is used to retrieve the closest body poses by searching the nearest neighbors, and the ground truth poses are averaged to predict the body pose in 3D. On the other hand, the deep learning approach uses a CNN model to estimate the human pose from the pressure maps. The experiments on a private dataset show a 12.20 mean absolute error (MAE) between the ground truth and the predicted joints for the hashing approach, while the CNN model achieves 8.00 MAE.

In \cite{obeidavi2022pose}, the study focuses on the use of a thermal camera image-based pose estimation model, both sustaining the privacy of the patient and robustness in occluded cases. The proposed Multi-Scale Stacked Hourglass (MSSHg) network is applied to enhance the processing of the images in different scales for pose estimation. The results demonstrate the effectiveness of the MSSHg network, achieving an accuracy of 96.8\% in the PCK0.2 on the SLP dataset~\cite{liu2022simultaneously}.

\subsection{Multimodal In-bed Pose Estimation}

Multimodal in-bed pose estimation studies employ multiple data sources, such as RGB, LWIR, pressure maps, and depth maps. In \cite{cao2022bed}, the authors design a multimodal pipeline to extract more informative features for training. However, they remove the non-privacy preserving RGB modality during inference time. To maintain the performance of the model, the authors propose a solution using a multimodal conditional variational autoencoder (MCVAE) in conjunction with the HRNet \cite{sun2019deep} model. This approach enables the reconstruction of features from missing modalities during test time, providing a self-supervised mechanism for learning. The results show that the proposed framework effectively predicts the joints from the available modality by achieving a 95.8 PCK0.5 score, compared to the method fusing all the modalities which achieved 95.6 PCK0.5 on the SLP dataset\cite{liu2022simultaneously}.

In \cite{dayarathna2023privacy}, the study utilizes multiple modalities such as RGB, depth maps, LWIR, and pressure maps. There are two primary objectives of this study: (1) The effective fusion of information from different modalities to enhance pose estimation and (2) the development of a framework capable of estimating in-bed poses with only the privacy-preserving modalities. Various fusion techniques are explored, including addition, concatenation, fusion via learnable parameters, and an end-to-end fully trainable approach coupled with a state-of-the-art pose estimation model. The study also introduces visible image reconstruction from privacy-preserving LWIR images using a conditional generative adversarial network (GAN). Experiments on the SLP dataset \cite{liu2022simultaneously} demonstrate the performance increase of the proposed fusion model compared to the recent unimodal approaches in the literature.

In \cite{yin2022multimodal}, the authors propose a pyramid scheme to effectively fuse different modalities, leveraging the knowledge captured by multimodal sensors. The depth and infrared images are initially fused to generate robust pose and shape estimations. Subsequently, pressure maps and RGB images are fused to refine the results, providing occlusion-invariant information for covered parts and accurate shape details for uncovered parts. An attention-based reconstruction module is proposed to generate uncovered modalities to mitigate occlusion effects. The pyramid fusion scheme and attention-based reconstruction module prove effective in complementing each modality, showcasing the contributions of depth, infrared, pressure map, and RGB data in the pose estimation process. According to the results, the fusion of modalities in multiple levels results in an MPJPE of 80.21 for thick-covered samples on the SLP dataset \cite{liu2022simultaneously}.

\section{Limitations and Potentials}
 The first limitation of the field can be given as the comparability of the current studies. Since different dataset portions are used during model evaluations in the papers, \textbf{a fair comparison between the approaches is not possible}. Furthermore, there is a lack of diverse datasets in terms of age and inclusion of people with disorders. In scenarios with a person having anatomical abnormalities or people from different ages, the proposed algorithms may not perform well due to limb sizes and joint locations from out-of-distribution. Therefore, the future \textbf{datasets should be sampled from a larger and diverse population}. 

Regarding the algorithm design, a potential that may improve the current state-of-the-art is to \textbf{include more modalities during model training}. The previous approaches generally utilize one or two modalities to estimate human poses, and only a few studies are exploring the use of more than two modalities. Besides, RGB images were mainly eliminated in the studies due to the visual attributes they include, which might breach individuals' privacy. However, the color and shape information obtained from the RGB images can be beneficial for the task. Moreover, some of the current approaches utilize RGB images during training time but eliminate them during inference time. To maintain the performance of the model, GAN models are used to generate the missing modalities during test time. However, these approaches can slow down the inference time by increasing the number of floating operations, and may not be suitable for real-time scenarios. Therefore, in the future, generating a new latent representation for the RGB images with lightweight models to benefit from the modality while preserving privacy, and the fusion of more than two modalities would be more appropriate in terms of speed, privacy, and accuracy.

\section{Conclusion}

In this review, we summarized the publicly available datasets and the in-bed human pose estimation approaches in the literature. It was shown that while the uncovered poses are relatively easier to estimate than the covered scenes, the fusion of different modalities both preserves the privacy of the patient and gives a comprehensive understanding of the person under the blanket, thus improving the accuracy. To eliminate the visual attributes of the patients, non-RGB modalities have been focused on more during the model development due to privacy concerns. It was observed that a common benchmark is necessary to evaluate proposed models since the current evaluation of the approaches is dependent on the selection of the training and test subsets. Finally, to evaluate the generalizability of the models, individuals with different anatomical characteristics and from a wider age range should be included in the datasets for future studies.

\section*{Acknowledgment}

This research work has been partially funded by GoodBrother Cost Action (CA19121) via a Short-Term Scientific Mission grant provided to Ziya Ata Yazıcı.

\newpage
\bibliographystyle{ieeetr}
\bibliography{bib}

\begin{thebibliography}{10}

\bibitem{wang2021deep}
J.~Wang, S.~Tan, X.~Zhen, S.~Xu, F.~Zheng, Z.~He, and L.~Shao, ``Deep 3d human pose estimation: A review,'' {\em Computer Vision and Image Understanding}, vol.~210, p.~103225, 2021.

\bibitem{jang2020etri}
J.~Jang, D.~Kim, C.~Park, M.~Jang, J.~Lee, and J.~Kim, ``Etri-activity3d: A large-scale rgb-d dataset for robots to recognize daily activities of the elderly,'' in {\em 2020 IEEE/RSJ International Conference on Intelligent Robots and Systems (IROS)}, pp.~10990--10997, IEEE, 2020.

\bibitem{groos2022towards}
D.~Groos, L.~Adde, R.~St{\o}en, H.~Ramampiaro, and E.~A. Ihlen, ``Towards human-level performance on automatic pose estimation of infant spontaneous movements,'' {\em Computerized Medical Imaging and Graphics}, vol.~95, p.~102012, 2022.

\bibitem{qiu2022pose}
Y.~Qiu, J.~Wang, Z.~Jin, H.~Chen, M.~Zhang, and L.~Guo, ``Pose-guided matching based on deep learning for assessing quality of action on rehabilitation training,'' {\em Biomedical Signal Processing and Control}, vol.~72, p.~103323, 2022.

\bibitem{srivastav2022unsupervised}
V.~Srivastav, A.~Gangi, and N.~Padoy, ``Unsupervised domain adaptation for clinician pose estimation and instance segmentation in the operating room,'' {\em Medical Image Analysis}, vol.~80, p.~102525, 2022.

\bibitem{clever20183d}
H.~M. Clever, A.~Kapusta, D.~Park, Z.~Erickson, Y.~Chitalia, and C.~C. Kemp, ``{3D Human Pose Estimation on a Configurable Bed from a Pressure Image},'' in {\em 2018 IEEE/RSJ International Conference on Intelligent Robots and Systems (IROS)}, pp.~54--61, IEEE, 2018.

\bibitem{liu2019bed}
S.~Liu, Y.~Yin, and S.~Ostadabbas, ``{In-Bed Pose Estimation: Deep Learning With Shallow Dataset},'' {\em IEEE Journal of Translational Engineering in Health and Medicine}, vol.~7, pp.~1--12, 2019.

\bibitem{carmona2022blanketset}
J.~Carmona, T.~Kar{\'a}csony, and J.~P.~S. Cunha, ``{BlanketSet--A Clinical Real Word Action Recognition and Qualitative Semi-synchronised MoCap Dataset},'' {\em arXiv preprint arXiv:2210.03600}, 2022.

\bibitem{liu2022simultaneously}
S.~Liu, X.~Huang, N.~Fu, C.~Li, Z.~Su, and S.~Ostadabbas, ``Simultaneously-collected multimodal lying pose dataset: Enabling in-bed human pose monitoring,'' {\em IEEE Transactions on Pattern Analysis and Machine Intelligence}, vol.~45, no.~1, pp.~1106--1118, 2022.

\bibitem{achilles2016patient}
F.~Achilles, A.-E. Ichim, H.~Coskun, F.~Tombari, S.~Noachtar, and N.~Navab, ``{ Patient MoCap: Human Pose Estimation Under Blanket Occlusion for Hospital Monitoring Applications},'' in {\em Medical Image Computing and Computer-Assisted Intervention--MICCAI 2016: 19th International Conference, Athens, Greece, October 17-21, 2016, Proceedings, Part I 19}, pp.~491--499, Springer, 2016.

\bibitem{liu2017vision}
S.~Liu and S.~Ostadabbas, ``{A Vision-Based System for In-Bed Posture Tracking},'' in {\em Proceedings of the IEEE International Conference on Computer Vision Workshops}, pp.~1373--1382, 2017.

\bibitem{davoodnia2022estimating}
V.~Davoodnia, S.~Ghorbani, and A.~Etemad, ``Estimating pose from pressure data for smart beds with deep image-based pose estimators,'' {\em Applied Intelligence}, vol.~52, no.~2, pp.~2119--2133, 2022.

\bibitem{bigalke2023anatomy}
A.~Bigalke, L.~Hansen, J.~Diesel, C.~Hennigs, P.~Rostalski, and M.~P. Heinrich, ``Anatomy-guided domain adaptation for 3d in-bed human pose estimation,'' {\em Medical Image Analysis}, vol.~89, p.~102887, 2023.

\bibitem{afham2022towards}
M.~Afham, U.~Haputhanthri, J.~Pradeepkumar, M.~Anandakumar, A.~De~Silva, and C.~U. Edussooriya, ``{Towards Accurate Cross-Domain in-Bed Human Pose Estimation},'' in {\em ICASSP 2022-2022 IEEE International Conference on Acoustics, Speech and Signal Processing (ICASSP)}, pp.~2664--2668, IEEE, 2022.

\bibitem{casas2019patient}
L.~Casas, N.~Navab, and S.~Demirci, ``Patient 3d body pose estimation from pressure imaging,'' {\em International Journal of Computer Assisted Radiology and Surgery}, vol.~14, pp.~517--524, 2019.

\bibitem{obeidavi2022pose}
S.~Obeidavi, M.~Gandomkar, and G.~Hirtz, ``{In-Pose Estimation of Covered and Uncovered Human Body from Thermal Camera Images Using Multi-Scale Stacked Hourglass (MSSHg) Network},'' in {\em 2022 16th International Conference on Signal-Image Technology \& Internet-Based Systems (SITIS)}, pp.~84--90, IEEE, 2022.

\bibitem{cao2022bed}
T.~Cao, M.~A. Armin, S.~Denman, L.~Petersson, and D.~Ahmedt-Aristizabal, ``{In-Bed Human Pose Estimation from Unseen and Privacy-Preserving Image Domains},'' in {\em 2022 IEEE 19th International Symposium on Biomedical Imaging (ISBI)}, pp.~1--5, IEEE, 2022.

\bibitem{dayarathna2023privacy}
T.~Dayarathna, T.~Muthukumarana, Y.~Rathnayaka, S.~Denman, C.~de~Silva, A.~Pemasiri, and D.~Ahmedt-Aristizabal, ``Privacy-preserving in-bed pose monitoring: A fusion and reconstruction study,'' {\em Expert Systems with Applications}, vol.~213, p.~119139, 2023.

\bibitem{yin2022multimodal}
Y.~Yin, J.~P. Robinson, and Y.~Fu, ``Multimodal in-bed pose and shape estimation under the blankets,'' in {\em Proceedings of the 30th ACM International Conference on Multimedia}, pp.~2411--2419, 2022.

\bibitem{zhu2017unpaired}
J.-Y. Zhu, T.~Park, P.~Isola, and A.~A. Efros, ``{Towards Accurate Cross-Domain in-Bed Human Pose Estimation},'' in {\em Proceedings of the IEEE International Conference on Computer Vision}, pp.~2223--2232, 2017.

\bibitem{sun2019deep}
K.~Sun, B.~Xiao, D.~Liu, and J.~Wang, ``{Deep High-Resolution Representation Learning for Human Pose Estimation},'' in {\em Proceedings of the IEEE/CVF Conference on Computer Vision and Pattern Recognition}, pp.~5693--5703, 2019.

\end{thebibliography}

\end{document}